\documentclass[11pt,a4paper]{article}
\usepackage[hyperref]{emnlp2018}
\usepackage{times}
\usepackage{latexsym}
\usepackage{url}

\usepackage{multirow}
\usepackage{booktabs}
\usepackage{caption}
\DeclareCaptionFont{10pt}{\fontsize{10pt}{12pt}\selectfont}
\usepackage{subcaption}
\usepackage[flushleft]{threeparttable}
\usepackage{todonotes}
\definecolor{DarkRed}{RGB}{130,25,0}

\newcommand{\missing}[1]{\textcolor{red}{#1}}
\newcommand{\ignore}[1]{}
\newcommand{\event}[1]{\textit{\textbf{#1}}}
\newcommand{\timex}[1]{\textit{\textbf{#1}}}
\newcommand{\idxevent}[3]{\event{e#1:\textrm{\color{#3}#2}}}
\newcommand{\refevent}[2]{\event{e\ref{#1}:#2}}
\newcommand{\idxtimex}[3]{\timex{t#1:\textrm{\color{#3}#2}}}

\newcommand{\rel}[1]{{\em #1}}
\newcommand{\type}[1]{\textsc{#1}}

\newcommand{\danchange}[1]{#1}
\newcommand{\best}[1]{\textbf{#1}}

\newcommand{\demo}{CogCompTime}
\newcounter{exctr}
\newcommand{\addexctr}{\refstepcounter{exctr}\theexctr}
\newcounter{eventCtr}
\newcommand{\addeventCtr}{\refstepcounter{eventCtr}\theeventCtr}
\newcounter{timexCtr}
\newcommand{\addtimexCtr}{\refstepcounter{timexCtr}\thetimexCtr}
\usepackage{balance}
\aclfinalcopy 

\title{ \vspace*{-0.5in}
	{{\small \hfill EMNLP'18}\\
		\vspace*{.25in}} CogCompTime: A Tool for Understanding Time in Natural Language}

\author{Qiang Ning,$^1$ Ben Zhou,$^1$ Zhili Feng,$^2$ Haoruo Peng,$^1$ Dan Roth$^{1,3}$ \\
Department of Computer Science\\
$^1$University of Illinois at Urbana-Champaign, Urbana, IL 61801, USA\\
$^2$University of Wisconsin-Madison, Madison, WI 53706, USA\\
$^3$University of Pennsylvania, Philadelphia, PA 19104, USA\\
{\tt \small \{qning2,xzhou45,hpeng7\}@illinois.edu,~zfeng49@cs.wisc.edu,~danroth@seas.upenn.edu}}

\date{}

\begin{document}
\maketitle

\begin{abstract}
Automatic \danchange{extraction}
of temporal information is important for
natural language understanding. It involves two basic tasks: (1) Understanding 
time expressions 
\danchange{that are mentioned} explicitly 
in text (e.g., \textit{February 27, 1998} or \textit{tomorrow}), and (2) Understanding temporal information 
\danchange{that is conveyed} implicitly  
via relations. 
This paper introduces \demo{}, \danchange{a system} that has these two important functionalities. It incorporates the most recent progress, achieves state-of-the-art performance, and is publicly available.\footnote{\url{http://cogcomp.org/page/publication_view/844}} 
We \danchange{believe}
that this demo \danchange{will}
provide valuable insight for temporal understanding and be useful for \danchange{multiple}
time-aware applications.
\end{abstract}

\section{Introduction}
Time is an important dimension when we describe the world because many facts are time-sensitive, e.g., one's place of residence, one's employment, or the progress of a conflict between countries.
\danchange{Consequently,}
many applications can benefit from temporal understanding in natural language, e.g., timeline construction \cite{DoLuRo12,MSAAEMRUK15}, clinical events analysis \cite{JindalRo13,BDSPV15}, question answering \cite{LCUMAP15}, and causality inference \cite{NingFeWuRo18}.

Temporal understanding from natural language requires two basic components \cite{VGSHKP07,VSCP10,ULADVP13}. 
The first, also known as the Timex component, 
requires extracting 
explicit time expressions in text (i.e., ``Timex'') and normalize them to a standard format. In Example~\ref{ex:timex 1}, the Timex is 
\timex{February 27, 1998} and its {\em normalized} form is 
`1998-02-27''. 
Note that normalization may also require a reference time for Timexes like ``tomorrow'', for which we need to know the document creation time (DCT).
In addition to \type{Date}, there are also other Timex types including \type{Time} (e.g., \timex{8 am}), \type{Duration} (e.g., \timex{3 years}), and \type{Set} (e.g., \timex{every Monday}).

Timexes carry temporal information {\em explicitly}, but temporal information can also be conveyed {\em implicitly} via temporal relations (i.e., ``TempRel''). In Example~\ref{ex:temprel 1}, there are two events: \refevent{ev:exploded}{exploded} and \refevent{ev:died}{died}. 
The text does not tell us when they happened, but we do know that there is a TempRel between them, i.e., \refevent{ev:exploded}{exploded} happened \rel{before} \refevent{ev:died}{died}.
The second basic component of temporal understanding is thus the TempRel component, which 
extracts TempRels automatically from text.
While the Timex component provides absolute time anchors, the TempRel component provides the relative order of events.
These two 
together
provide a 
complete picture of the temporal dimension of a story, so they are naturally the most important building blocks towards temporal understanding.

\begin{table}[h!]
	\centering\small
	\begin{tabular}{|p{7cm}|}
		\hline
		\textbf{Example~\addexctr\label{ex:timex 1}:}
		Presidents Leonid Kuchma of Ukraine and Boris Yeltsin of Russia signed an economic cooperation plan on (\idxtimex{\addtimexCtr\label{tx:19980227}}{February 27, 1998}{black}).\\
		\hline
		\textbf{Example~\addexctr\label{ex:temprel 1}:}
		A car (\idxevent{\addeventCtr\label{ev:exploded}}{exploded}{black}) in the middle of a group of men playing volleyball. More than 10 people have (\idxevent{\addeventCtr\label{ev:died}}{died}{black}), police said.\\
		\hline
	\end{tabular}
\end{table}

\begin{figure*}[h!]
	\centering
	\includegraphics[width=.93\textwidth]{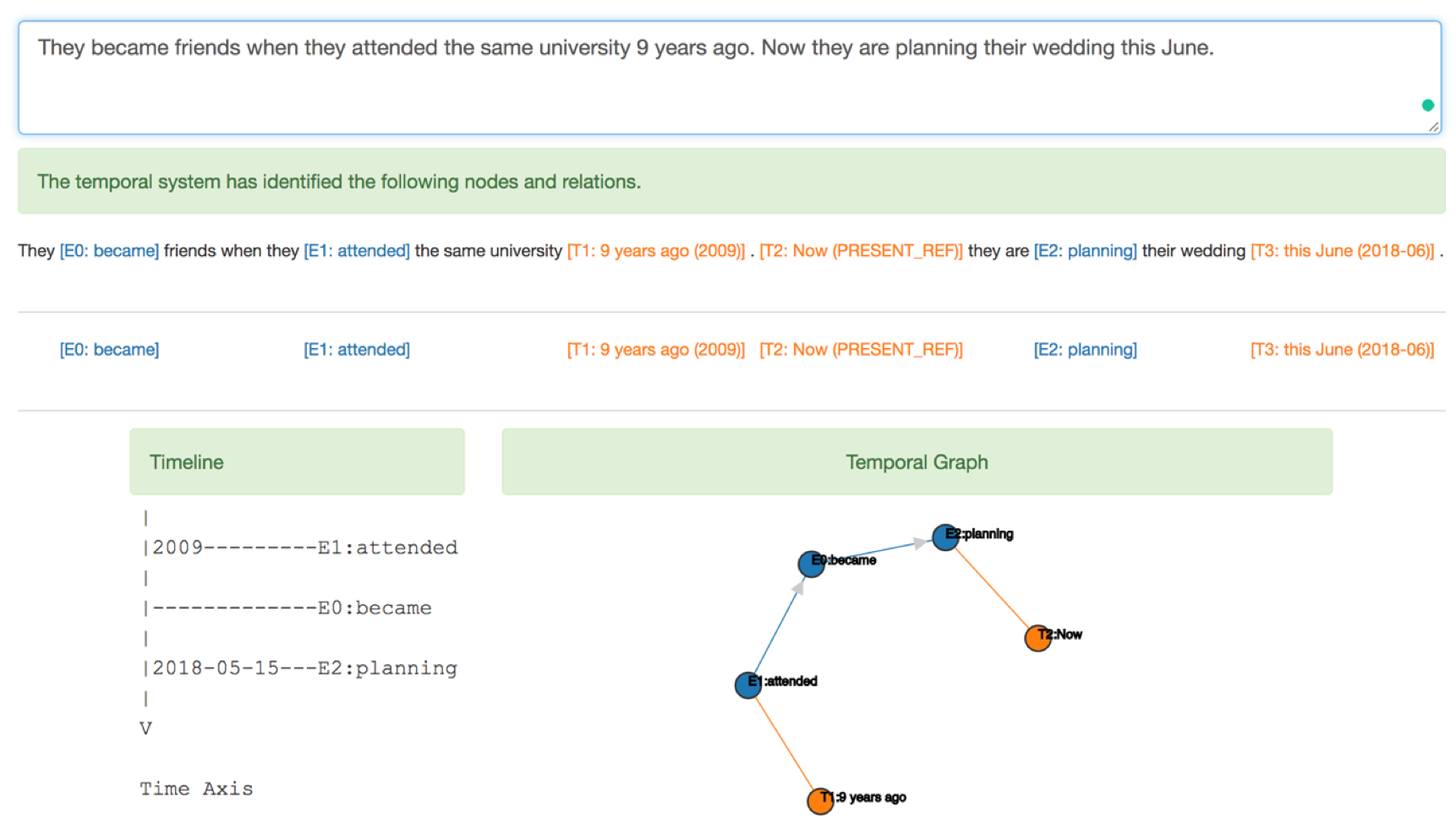}
	\caption{\textbf{A snapshot of the interface of \demo{}.} From top to bottom: Input box, event and Timex highlight, and two visualizations (timeline and graph). The document creation time was chosen to be 2018-05-15.}
	\label{fig:demo}
\end{figure*}

In this paper, we present \demo{} (see Fig.~\ref{fig:demo}), a tool with both the Timex and TempRel components, which are conceptually built on \citet{ZhaoDoRo12} and \citet{NingFeRo17}, respectively. \demo{} is a new implementation that integrates both components and also incorporates the most recent advances in this area \cite{NingFeWuRo18,NingWuPeRo18,NingWuRo18}.
Two highlights are: First, \demo{} achieves comparable performance to state-of-the-art Timex systems, but is almost two times faster than the second fastest, HeidelTime \cite{StrotgenGe10}.
Second, \demo{} improves the performance of the TempRel component by a large margin, from the literature's $F_1\approx 50$ \cite{ULADVP13} to $F_1\approx 70$ (see details in Sec.~\ref{sec:result}).
Given these two contributions, we believe that \demo{} is a good demonstration of the state-of-the-art in temporal understanding.
In addition, since \demo{} is publicly available, it will provide easy access to users working on time-aware applications, as well as
valuable insight to researchers seeking further improvements.

We briefly review the literature and explain in detail the processing pipeline of \demo{} in Sec.~\ref{sec:system}: the Timex component, the Event Extraction component, and the TempRel component.
Following that, we provide a benchmark evaluation in Sec.~\ref{sec:result} on the TempEval3 and the MATRES datasets \cite{ULADVP13,NingWuRo18}.
Finally, we point out directions for future work and conclude this paper.
\begin{figure*}[htbp!]
	\centering
	\includegraphics[width=.85\textwidth]{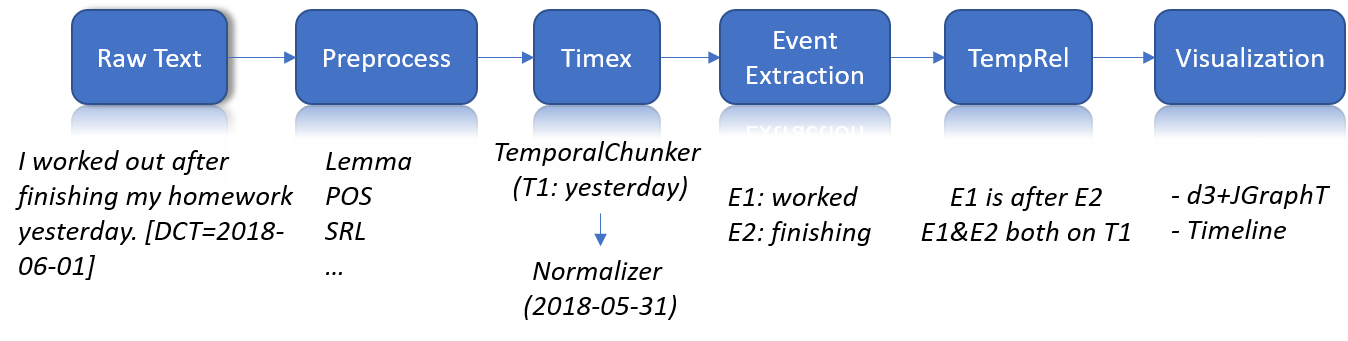}
	\caption{\small\textbf{System pipeline of \demo{}}: It preprocesses raw text input using CogCompNLP and then applies Timex, Event Extraction, and TempRel components sequentially, with two user-friendly visualizations (i.e., graph-type visualization and timeline-type visualization) provided at the end.}
	\label{fig:pipeline}
\end{figure*}

\section{System}
\label{sec:system}
The system pipeline of \demo{} is shown in Fig.~\ref{fig:pipeline}:
It takes raw text as input and uses CogCompNLP \cite{KSZRCS18} to extract features such as lemmas, part-of-speech (POS) tags, and semantic role labelings (SRL).
Then \demo{} sequentially applies the Timex component, the event extraction component, and the TempRel component. Finally, both a graph visualization and a timeline visualization are provided for the users. In the following, we will explain these main modules in detail.

\subsection{Timex Component}
Existing work on Timex extraction and normalization falls into two categories: rule-based and learning-based. Rule-based systems use regular expressions to extract Timex in text and then deterministic rules to normalize them (e.g., HeidelTime \cite{StrotgenGe10} and SUTime \cite{ChangMa12}). Learning-based systems use classification models to chunk out Timexes in text and normalize them based on grammar parsing (e.g., UWTime \cite{LADZ14} and ParsingTime \cite{AngeliMaJu12}).
\demo{} adopts a mixed strategy: we use machine learning in the Timex extraction step and rule parsing in the normalization step. This mixed strategy, while maintaining a state-of-the-art performance, significantly improves the computational efficiency of the Timex component, as we show in Sec.~\ref{sec:result}.

Technically, the Timex extraction step can be formulated as a generic text chunking problem and the standard B(egin), I(nside), and O(utside) labeling scheme can be used.
\demo{} proposes TemporalChunker, by retraining Illinois-Chunker \cite{PunyakanokRo01} on top of the Timex chunk annotations provided by the TempEval3 workshop \cite{ULADVP13}.
Here a machine learning based extraction algorithm significantly improves the computational efficiency by quickly sifting out impossible text chunks, as compared to regular expression matching, which has to check every substring of text against regular expressions and is often slow. However, we do admit that learning-based extraction handles corner cases not as well as rule-based systems because of the limited training examples.

After Timexes are extracted, we apply rules to normalize them. We think rule-based methods are generally more natural for normalization: On one hand, the desired formats of various types of Timexes are already defined as rules by corresponding annotation guidelines; on the other hand, the intermediate steps of how one Timex is normalized are not annotated in any existing datasets (it is inherently hard to do so), so learning-based methods usually have to introduce latent variables and need more training instances as a result.
Therefore, we have adopted a rule-based normalization method.
However, we admit that an obvious drawback is that the rule set needs to be redesigned for every single language.

\subsection{Event Extraction Component}

Event extraction is closely related to how events are defined.
Generally speaking, an event is considered to be an action associated with the corresponding participants.
In the context of temporal understanding, events are usually represented by their head verb token, so unlike the generic chunking problem in Timex extraction, event extraction can be formulated as a classification problem for each token. Specifically, \demo{} only considers those {\em main-axis} events, so event extraction is simply a binary classification problem (i.e., whether or not a token is a main-axis event or not).
As defined by the MATRES annotation scheme \cite{NingWuRo18}, main-axis events are those events that form the primary timeline of a story and approximately 60\%-70\% of the verbs are on the main-axis in MATRES.
We extract lemmas and POS tags within a fixed window, SRL, and prepositional phrase head, and train a sparse averaged perceptron for event extraction.

\subsection{TempRel Component}
Temporal relations can be generally modeled by a graph (called temporal graph), where the nodes represent events and Timexes, and the edges represent TempRels.
With all the nodes extracted (by previous steps), the TempRel component is to make predictions on the labels of those edges. In this paper, the label set for Event-Event TempRels  is \rel{before}, \rel{after}, \rel{equal}, and \rel{vague} and for Event-Timex TempRels is \rel{equal} and \rel{not-equal}.\footnote{The simplification of Event-Timex label set is due to our observation that other labels have very low accuracies. As a demo paper, we have chosen not to use them. However, we think it is interesting and worth further investigation.}
State-of-the-art methods include, e.g., ClearTK \cite{Bethard13}, CAEVO \cite{CCMB14}, and \citet{NingFeRo17}.
The TempRel task is known to be very difficult.
\citet{NingWuRo18} attributes the difficulty partly to the low inter-annotator agreement (IAA) of existing TempRel datasets and proposes a new Multi-Axis Temporal RElations dataset of Start-points (MATRES) with significantly improved IAA, so for the TempRel task, we have chosen MATRES as the benchmark in this paper.\footnote{Specifically, we only need to replace the TempRel annotations in TempEval3 by the new annotations in MATRES.}

\begin{figure}[htbp!]
	\centering
	\includegraphics[width=.45\textwidth]{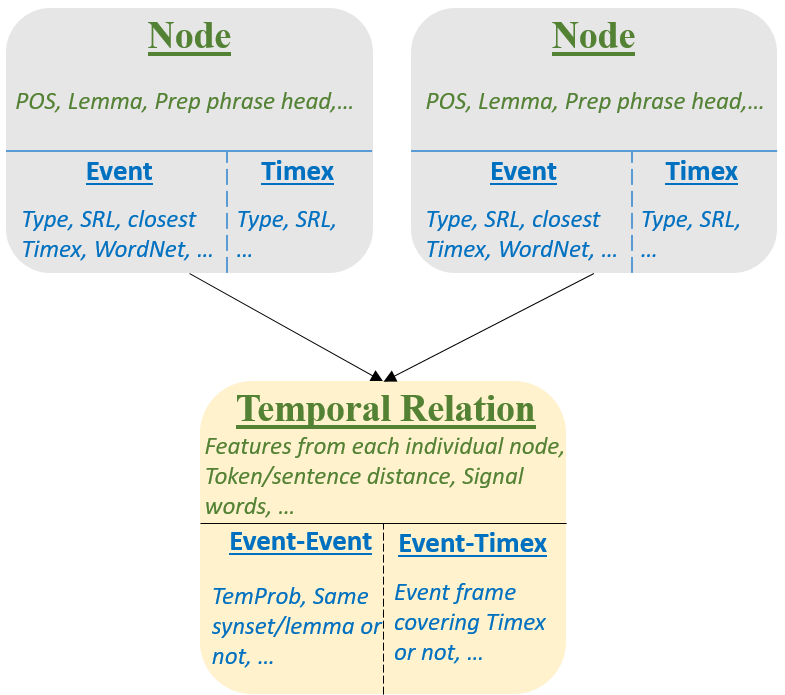}
	\caption{\small\textbf{The primary features used in the TempRel component} (also standard features used in the literature). Since there are two types of nodes (i.e., event and Timex), and two types of TempRels (i.e., Event-Event and Event-Timex), we put the common features above and split specific feature sets below. Conjunctive features are not listed exhaustively here.}
	\label{fig:features}
\end{figure}

We also incorporate the recent progress of \citet{NingFeRo17,NingFeWuRo18,NingWuPeRo18}.
The feature set used for TempRel is shown in Fig.~\ref{fig:features}, which contains features derived individually from each node and jointly from a node pair. Since a node can be either an event or a Timex, an edge can also be either an Event-Event edge or an Event-Timex edge and the features have to vary a bit, as detailed by Fig.~\ref{fig:features}.
Note that for Event-Event edges, we incorporate features from TemProb,\footnote{\url{http://cogcomp.org/page/resource_view/114}} which encodes prior knowledge of typical temporal orders of events \cite{NingWuPeRo18}.
With these features, we also adopt the constraint-driven learning algorithm for TempRel classification proposed in \citet{NingFeRo17} with sparse averaged perceptron.
Then our TempRel component assigns local prediction scores (i.e., soft-max scores) to each edge and solves an integer linear programming (ILP) problem via Gurobi \cite{Gurobi15} to achieve globally consistent temporal graphs (please refer to \citet{NingFeWuRo18} for details).
\demo{} is a unique package so far that incorporates all the recent progress.

\subsection{Visualization}
As shown in Fig.~\ref{fig:demo}, we highlight the extracted Timexes and events in the text. Specifically for Timexes, we also annotate their normalized values along with their chunks.
We provide two forms of visualization for the extracted TempRels. Since TempRels can be naturally modeled by a graph,  a graph visualization is an obvious choice and we use d3 (\url{https://d3js.org/}) in \demo{}.
Additionally, we provide a more compact visualization to those graphs via timeline construction. Since a graph is only partially ordered (as opposed to a timeline which is fully ordered), we resort to the appearance order of events in timeline construction when the temporal order is \rel{vague} according to its graph.

\section{Benchmark Experiment}
\label{sec:result}
\begin{table*}[htbp!]
	\centering\small
	\begin{tabular}{c|c|c|c|c|c|c}
		\hline
		\multirow{2}{*}{Timex Systems} & \multicolumn{3}{c|}{Extraction} & Normalization & End-to-end & Runtime\\
		\cline{2-7}
		&	P	&	R	&	$F_1$	&	Accuracy	&	$F_1$	&	Seconds\\
		\hline
		HeidelTime \cite{StrotgenGe10}	&	84.0	&	79.7	&	81.8	&	78.1$^\dagger$ &	78.1	&	18\\
		SUTime \cite{ChangMa12}		&	80.0	&	81.1	&	80.6	&	69.8$^\dagger$&	69.8	&	16\\
		UWTime \cite{LADZ14}		&	\best{86.7}	&	80.4	&	83.5	&	84.4	&	\best{82.7}	&	400\\
		\demo{}		&	86.5	&	\best{83.3}	&	\best{84.9}	&	\best{84.7}	&	76.8	&	\best{7}\\
		\hline
	\end{tabular}
	\caption{\small\textbf{Performance of our Timex component} compared with state-of-the-art systems on a benchmark dataset, the Platinum dataset from the TempEval3 workshop \cite{ULADVP13}. The ``extraction'' and ``normalization'' columns are the two intermediate steps. ``Normalization'' was evaluated given gold extraction, while ``end-to-end'' means system extraction was used. Runtimes were evaluated under the same setup.}
	
	\begin{tablenotes}\small
		\item $^\dagger$HeidelTime and SUTime have no clear-cut between extraction and normalization, so even if gold Timex chunks are fed in, their extraction step cannot be easily skipped.
	\end{tablenotes}
	
	\label{tab:timex}
\end{table*}

We used the dataset provided by the TempEval3 workshop, with the original train/test split in our experiment: TimeBank and AQUAINT were for training (256 articles), and Platinum was for testing (20 articles).
Note that we replaced the TempRel annotations in the original TempEval3 datasets by MATRES due to its higher IAA. 
In the Timex component, TemporalChunker by default takes 10\% of the train set as the development set, and in other components, 5-fold cross-validation was used for parameter tuning.

Table~\ref{tab:timex} evaluates the Timex component of \demo{}, comparing with state-of-the-art systems.
The ``normalization'' and ``end-to-end'' columns were evaluated based on gold Timex extraction and system Timex extraction, respectively.
The fact that \demo{} had the best extraction $F_1$ and normalization accuracy but not the best end-to-end performance is due to our mixed strategy: Timexes extracted by our learning-based TemporalChunker sometimes cannot be normalized correctly by our rule-based normalizer. This phenomenon is relatively more severe in \demo{} comparing to systems that are consistently rule-based or learning-based in both extraction and normalization. However, the computational efficiency is improved significantly by reducing the runtime of the second fastest, HeidelTime, by more than 50\%.

Table~\ref{tab:temprel} shows the performance of the Event Extraction and TempRel components. We also copied the Timex extraction performance from Table~\ref{tab:timex}.
Note that \demo{} only extracts those main-axis events as defined by MATRES.
Since \citet{NingWuRo18} did not propose an event extraction method, Table~\ref{tab:temprel} is in fact the first reported performance of event extraction on MATRES and as we see, both the precision and recall are better than those numbers reported in TempEval3. Note that since \demo{} works on different annotations, this does not indicate that our event extraction algorithm is better than those participants in TempEval3; instead, this indicates that the event extraction problem in MATRES is a better-defined machine learning task.

The performance of TempRel extraction is further evaluated in Table~\ref{tab:temprel}, both when the gold event and Timex extraction is used and when system extraction is used.
As for Event-Event TempRels, we also introduce a new relaxed metric\footnote{This relaxed metric does not apply to Event-Timex TempRels since the label set is only \rel{equal} and \rel{not-equal}, .}, where predictions of \rel{before}/\rel{after} are not penalized when the gold label is \rel{vague}. This is based on the definition of \rel{vague} in MATRES, i.e., to assign \rel{vague} labels when either \rel{before} or \rel{after} is reasonable. We think this relaxed metric is more suitable when creating timelines from temporal graphs, where an order must be picked anyhow when two events have a \rel{vague} relation.
When system extraction was used, the TempRel performance saw a large drop. However, the performance here, although it is perhaps still not sufficiently good for some applications, is already a significant step forward in temporal understanding. 
As a reference point, the best system in TempEval3, ClearTK \cite{Bethard13}, had P=34.08, R=28.40, $F_1$=30.98 (using system extraction) and P=37.32, R=35.25, $F_1$=36.26 (using gold extraction). Again, given the dataset difference, these numbers are not directly comparable, but it indicates that the MATRES dataset used here probably has the TempRel task better defined and we hope this demo paper will be a good showcase of the new state-of-the-art.

\begin{table}[htbp!]
	\centering\small
	\begin{tabular}{l|c|c|c}
		\hline
			&	P	&	R	&	$F_1$\\
		\hline
		Event Extraction	&	83.5	&	87.0	&	85.2\\
		Timex Extraction	&	86.5	&	83.3	&	84.9\\
		\hline
		\multicolumn{4}{c}{\textbf{Gold Extraction}}\\
		\hline
		Event-Event	&	61.6	&	70.9	&	65.9\\
		Event-Event (Relaxed)	
					& 	75.2	&	74.8	&	75.0\\
		Event-Timex	&	84.6	&	84.6	&	84.6\\
		\hline
		\multicolumn{4}{c}{\textbf{System Extraction}}\\
		\hline
		Event-Event	& 	48.4	&	58.0	&	52.8\\
		Event-Event (Relaxed)	
					& 	75.6	&	61.8	&	68.0\\
		Event-Timex	& 	79.5	&	61.1	&	69.0\\
		\hline
	\end{tabular}
	\caption{\small\textbf{Performance of the Event/Timex Extraction and TempRel components} when gold/system extraction is used. The relaxed metric does not penalize the system if a \rel{before}/\rel{after} prediction is made on a \rel{vague} relation. Please also refer to the text about this metric.}
	\label{tab:temprel}
\end{table}

\ignore{
\begin{table*}[htbp!]
	\centering
	\begin{tabular}{c|c|c|c|c|c|c}
		\hline
		\multirow{2}{*}{Event-Event TempRels}	&	\multicolumn{3}{c|}{Conventional Metric}	&	\multicolumn{3}{c}{Awareness Metric}\\
		\cline{2-7}
		&	P	&	R	&	$F_1$	&	P	&	R	&	$F_1$\\
		\hline
		Relation only	&	64	&	74	&	68	& 55.0	& 64.8	& 59.5\\
		Relation only (Relaxed)	&	75.2	&	74.8	&	75.0	& 71.9	& 66.2	& 68.9\\
		\hline
		End-to-end	&	49&	58&53	&43	&52	&47\\
		End-to-end (Relaxed)	&75.6	&61.8	&68.0	&72.1	&67.5	&69.7\\
		\hline
	\end{tabular}
	\caption{\missing{Event-Event TempRels}}
	\label{tab:temprel2}
\end{table*}
}
\section{Future Work}
We plan to further improve \demo{} in the following directions.
First, the MATRES dataset \cite{NingWuRo18} only considers verb events, but nominal events are also very common and important, so we plan to incorporate nominal event extraction and corresponding TempRel extraction.
Second, \demo{} currently does not incorporate \danchange{an} event coreference component. Since coreference is important for bridging long-distance event pairs, it is a desirable feature. We can adopt existing event coreferencing techniques such as \citet{PengSoRo16} in the next step.
Third, \demo{} currently only works on the main-axis events as defined in MATRES. How to incorporate other axes, e.g., intention axis, opinion axis, and hypothesis axis, requires further investigation.

\section{Conclusion}
This paper presents \demo{}, a new package that, given
raw text, (1) extracts time expressions (Timex) and normalizes them to a standard format, and (2) extracts events on the main time axis of a story and the temporal relations between events and Timexes.
\demo{} takes advantage of many recent advances and achieves state-of-the-art performance in both tasks.
We think this demo will be interesting for a broad audience because it is useful not only for identifying the shortcomings of existing methods, but also for applications that depend on the temporal understanding of natural language text.

\section*{Acknowledgements}
This research is supported in part by a grant from the Allen Institute for Artificial Intelligence (allenai.org); the IBM-ILLINOIS Center for Cognitive Computing Systems Research (C3SR) - a research collaboration as part of the IBM AI Horizons Network; and by the Army Research Laboratory (ARL) and was accomplished under Cooperative Agreement Number W911NF-09-2-0053 (the ARL Network Science CTA).
The views and conclusions contained in this document are those of the authors and should not be interpreted as representing the official policies, either expressed or implied, of the Army Research Laboratory or the U.S. Government. The U.S. Government is authorized to reproduce and distribute reprints for Government purposes notwithstanding any copyright notation here on.

\bibliography{emnlp2018,ccg-long,cited-long}
\bibliographystyle{acl_natbib_nourl}


\end{document}